\documentclass[runningheads]{llncs}
\usepackage[T1]{fontenc}
\usepackage{graphicx}
\usepackage{booktabs}
\usepackage[misc]{ifsym}
\newcommand{\corr}{(\Letter)}

\usepackage{xcolor, colortbl}
\usepackage{tikz}      
\usepackage{float}
\usepackage{cite}
\usepackage{url}
\usepackage{amsmath}
\usepackage{tikz}
\usetikzlibrary{arrows.meta, positioning, shapes.misc}
\usepackage{mwe}

\begin{document}

\title{Objective Mispricing Detection for Shortlisting Undervalued Football Players via Market Dynamics and News Signals}

\titlerunning{Shortlisting Undervalued Football Players via Market\,+\,News}

\author{Chinenye Ekene Omejieke \corr \inst{1} \and
Shuyao Chen\inst{2} \and
Xia Cui}

\authorrunning{C.E. Omejieke et al.}

 \institute{School of Computing and Mathematics, Manchester Metropolitan University, Manchester, United Kingdom \email{chinenyevans@gmail.com}, \email{x.cui@mmu.ac.uk}
 \and
 Typewind Ltd.,  London, United Kingdom \email{contact@typewind.co.uk}}

\maketitle              

\begin{abstract}
We present a practical, reproducible framework for identifying \emph{undervalued} football players grounded in \emph{objective mispricing}. Instead of relying on subjective expert labels, we estimate an expected market value from structured data (historical market dynamics, biographical and contract features, transfer history) and compare it to the observed valuation to define mispricing. We then assess whether news-derived Natural Language Processing (NLP) features (i.e., sentiment statistics and semantic embeddings from football articles) \emph{complement} market signals for shortlisting undervalued players.

Using a chronological (leakage-aware) evaluation, gradient-boosted regression explains a large share of the variance in log-transformed market value. For undervaluation \emph{shortlisting}, ROC-AUC-based ablations show that market dynamics are the primary signal, while NLP features provide consistent, secondary gains that improve robustness and interpretability. SHAP analyses suggest the dominance of market trends and age, with news-derived volatility cues amplifying signals in high-uncertainty regimes. The proposed pipeline is designed for decision support in scouting workflows, emphasizing ranking/shortlisting over hard classification thresholds, and includes a concise reproducibility and ethics statement.

\end{abstract}

\section{Introduction}
Valuing professional football players underpins transfer decisions, contract negotiations, and long-term squad planning~\cite{follert2024decision}. Yet market valuations remain noisy proxies of true sporting and economic value due to incomplete information, behavioural biases, contractual frictions, and media narratives~\cite{franck2012talent,herm2014crowd}. Identifying \emph{undervalued} players (i.e., those whose market prices fall below their underlying worth) offers clear competitive and financial benefits.

Most valuation models rely on structured indicators (performance, age trajectories, fees) and often assume near-efficient markets~\cite{franceschi2024determinants}. However, documented transfer-market inefficiencies arise from asymmetric information and delayed reactions~\cite{van2011valuation}. Unstructured news can affect perceptions through narratives about form, injuries and roles; analogous effects are well known in financial markets~\cite{tetlock2007giving, bollen2011twitter}. Integrating such text with market dynamics is therefore attractive but methodologically challenging; prior sports work often uses coarse sentiment and rarely quantifies the incremental value of NLP relative to structured data~\cite{waskita2025role, schumaker2016predicting}.

We propose an applied, leakage-aware framework for \textit{objective mispricing} detection: we estimate an expected value without using the current valuation as an input, compute mispricing as expected vs.\ observed, and label the upper quantile as \textit{undervalued}. This provides a reproducible and domain-grounded target aligned with scouting practice. We then study whether news-derived NLP features complement market dynamics for shortlisting undervalued players (ranking), rather than aiming at binary classification with a fixed threshold.

The primary contributions of this study are:
\begin{itemize}
    \item A reproducible mispricing-based definition of undervaluation that avoids subjective expert judgements.

    \item A practical integration of market dynamics, contract/transfer context and news-derived sentiment/embeddings to shortlist undervalued players.
    
    \item A systematic ablation demonstrating that NLP features provide consistent \textit{secondary} gains atop market dynamics, improving robustness and interpretability.
    
    \item An interpretable SHAP explanations articulate why candidates are surfaced, facilitating analyst trust and adoption.

\end{itemize}

\section{Related Work}
Research on football player valuation spans economics, sports analytics, and machine learning, reflecting the sizable financial stakes associated with transfer markets and contract negotiations. Early work modeled player value using classical econometric and hedonic pricing frameworks, where market valuations were expressed as functions of observable performance attributes such as age, position, playing time, and productivity. Megia~\cite{megia2025football} demonstrated that hedonic models can explain part of the variance in market value but remain sensitive to institutional and behavioral influences, indicating that observable performance variables alone are insufficient.

Subsequent studies incorporated additional contextual factors. Franck and Nüesch~\cite{franck2012talent} showed that popularity-related signals, including media exposure and fan attention, systematically inflate valuations beyond on-field contributions. Herm \textit{et al.}~\cite{herm2014crowd} found that both crowd-sourced and expert assessments correlate with market prices but leave a substantial portion unexplained, suggesting that football markets are only partially efficient. Such findings indicate that latent player quality and market perception are imperfectly captured by structured indicators.

Machine learning approaches have increasingly been adopted to model non-linear interactions in valuation. Behravan \textit{et al.}~\cite{behravan2021novel} applied ensemble learning methods and revealed systematic mispricing related to age, league exposure, and contract structure. Decroos \textit{et al.}~\cite{decroos2019actions} proposed action-based valuation models that quantify on-ball contributions with high granularity, though such work depends on proprietary event-level data that may not be generally available. Complementing micro-level perspectives, macroeconomic analyses have examined transfer-market inefficiencies at an institutional scale: Jacobsen~\cite{jacobsen2023managing} highlighted how governance constraints, bargaining imbalances, and information asymmetries distort transfer outcomes and create prolonged deviations from efficient pricing. Collectively, these strands motivate the search for data-driven frameworks capable of identifying undervalued players before markets adjust.

Unstructured data has recently emerged as an additional source of signal. Reviews by Schumaker \textit{et al.}~\cite{schumaker2016predicting} and Zadeh~\cite{h2021quantifying} highlight the use of sentiment and textual cues in match prediction, engagement analysis, and performance assessment. Football news frequently encodes soft information about injuries, tactical roles, psychological factors, or transfer intentions (i.e., signals that may influence perceived value before they materialize in performance statistics). Parallel evidence in financial economics supports the role of media in shaping asset prices: Tetlock~\cite{tetlock2007giving} showed that negative news sentiment depresses equity markets, while Bollen \textit{et al.}~\cite{bollen2011twitter} found that collective mood extracted from social media predicts market movements. Such findings strengthen the theoretical justification for modeling football-player valuation using media-derived sentiment and semantic cues.

However, most sports analytics work using text relies on lexicon-based sentiment or shallow machine learning models that fail to capture the contextual richness of sports journalism. Advances in natural language processing, particularly transformer architectures~\cite{vaswani2017attention}, enable deep semantic embeddings that encode bidirectional context. Large-scale models such as BERT~\cite{devlin2019bert} and DistilBERT~\cite{sanh2019distilbert} provide dense representations that move beyond keyword frequencies and can capture subtle narrative signals (e.g., form trajectories, evolving roles, injury uncertainty).

This paper extends prior work by integrating both transformer-based sentiment and dense semantic embeddings into a unified valuation and shortlisting framework. Unlike earlier studies, we rigorously quantify the \textit{incremental} contribution of textual features relative to structured market and biographical data through systematic ablations, and we define undervaluation objectively using mispricing between expected and observed market values. This positions our work at the intersection of valuation modeling, market inefficiency analysis, and modern NLP-driven sports analytics.

\section{Methodology}
This section describes the construction of the multi-modal dataset (Section~\ref{sec:data}), the preprocessing pipeline (Section~\ref{sec:preprocessing}), the extraction of news-derived features (Section~\ref{sec:feat-extract}), and the modelling framework for estimating expected market values and identifying undervalued players via an objective mispricing formulation (Section~\ref{sec:mispricing}). All components are designed to ensure strict temporal integrity, reproducibility, and applicability within real-world scouting workflows.

\subsection{Data and Features}
\label{sec:data}

Structured data were collected directly from Transfermarkt~\footnote{\url{https://www.transfermarkt.com/}}, including player biographical attributes (age, height, nationality), contract information (start date, expiry date, remaining duration), and longitudinal club affiliations. Weekly market-value snapshots from 2015--2024 provide a detailed view of valuation trajectories from which market-dynamics features were derived. Transfer histories were obtained in parallel, containing transfer dates, origin and destination clubs, declared fees, and transfer categories. Together, these sources provide long-term signals of economic stability, mobility, and performance-related valuation changes.

To complement the structured dataset, we constructed a multi-source corpus of football news. Articles were retrieved from multiple outlets including The Guardian, Sky Sports, BBC Sports, The Athletic, ESPN FC, and BeSoccer via the NewsAPI platform\footnote{\url{https://newsapi.org/}}, using player names and identifiers as search keys. To expand coverage beyond indexed sources, we additionally scraped publicly accessible football news websites. All articles were stored with full text and publication timestamps, enabling strict chronological alignment with market-value data. The combined dataset includes 19,982 players, 70,627 news articles, and 280,334 valuation snapshots (Table~\ref{tab:dataset_overview}).

\begin{table}[H]
\centering
\caption{Summary statistics of the integrated dataset (07/05/2015--23/01/2024).}
\label{tab:dataset_overview}
\begin{tabular}{lrl}
\toprule
\textbf{Item} & \textbf{Value} & \textbf{Description} \\
\midrule
Unique Players & 19,982 & Structured player records from Transfermarkt \\
News Articles & 70,627 & Multi-source football news mentioning players \\
Market Value Records & 280,334 & Weekly time-stamped valuations \\
Avg.\ Articles/Player & 44.20 & Mean article count per player \\
Temporal Range & 2015--2024 & Shared coverage period \\
\bottomrule
\end{tabular}
\end{table}

\subsection{Data Preprocessing}
\label{sec:preprocessing}
Raw data were processed at two levels: player market dynamics and news articles. From market-value time series, we derived player-level features including average valuation, volatility, maximum historical value, and a least-squares trend coefficient to capture career trajectory. Contract durations and transfer frequencies were also computed to reflect economic and career stability. Player names in news articles were normalized and mapped to unique entity identifiers using string matching and resolution heuristics, ensuring consistent cross-source linkage.

News articles were preprocessed using \texttt{spaCy} and regular-expression cleaning. This involved lowercasing, removal of HTML markup and boilerplate text, punctuation normalization, and sentence segmentation. Articles with insufficient content (e.g., fewer than 3 words) were discarded to maintain minimum information density, as research indicates that linguistic metrics become unstable and highly dependent on length in short-form texts~\cite{mccarthy2007vocd}. All article timestamps were preserved, and every transformation was applied to prevent information from the future influencing past data, adhering to best practices to avoid temporal leakage~\cite{mammadov2024optimizing}. This chronological discipline ensures our evaluation reflects real-world deployment.

\subsection{News Features}\label{sec:feat-extract}

\subsubsection{Sentiment Analysis.}
To capture tonal context, we extracted sentiment signals from each player's associated news articles. Representations were initially obtained using the Hugging Face \textit{BERT-base-uncased} transformer~\cite{devlin2019bert}. For sentiment polarity, we employed a \textit{DistilBERT} classifier~\cite{sanh2019distilbert} due to its computational efficiency (40\% smaller and 60\% faster than BERT while retaining 97\% of its language understanding) which was critical for processing our high-volume dataset. Sentiment scores were normalized to \([-1, +1]\). For each player \(i\), we aggregated mean sentiment \(\mu_i\) and sentiment volatility \(\sigma_i\):
\begin{equation}
\mu_i = \frac{1}{n_i}\sum_{j=1}^{n_i}s_{ij}, \qquad 
\sigma_i^2 = \frac{1}{n_i-1}\sum_{j=1}^{n_i}(s_{ij}-\mu_i)^2.
\end{equation}
Such signals have been shown to influence market-based outcomes~\cite{davidovic2025news}.

\subsubsection{Semantic Embeddings.}
To capture richer narratives—form, injuries, transfer talks—we encoded each article using a pre-trained sentence-level transformer. Tokenization employed the model's native WordPiece tokenizer with a maximum sequence length of 512. The \texttt{[CLS]} token representation was used as a holistic summary embedding. Player-level representations were obtained by mean pooling:
\begin{equation}
\bar{\mathbf{e}}_i = \frac{1}{n_i}\sum_{j=1}^{n_i}\mathbf{e}_{ij}.
\end{equation}
These contextual embeddings outperform bag-of-words approaches in capturing nuanced semantics~\cite{xia2023mole, reimers2019sentence}. To mitigate dimensionality and multicollinearity, we applied the Principal Component Analysis (PCA) to the 768-dimensional embeddings, retaining 95\% of cumulative variance. The PCA was fitted exclusively on the training set to ensure data isolation. The resulting low-dimensional features were concatenated with the structured market features of Section~\ref{sec:preprocessing}.

\subsection{Objective Mispricing and Shortlisting}\label{sec:mispricing}
\subsubsection{Expected Market Value Estimation.}
Market values are right-skewed; without transformation, a regressor would over-prioritize high-value players. To stabilize variance and linearize exponential growth, we applied a log transform. An expected value model was trained on all feature groups (excluding the current value) using XGBoost, linear regression, and TabNet. For player \(i\) with observed value \(V_i\):
\begin{equation}
y_i = \log(1 + V_i), \quad \hat{y}_i = f(\mathbf{x}_i,\mathbf{z}_i), \quad \hat{V}_i = \exp(\hat{y}_i) - 1.
\end{equation}
The addition of 1 ensures stability for zero-value entries.

\subsubsection{Mispricing and Undervaluation Label.}
Mispricing is the log-difference between expected and observed values:
\begin{equation}
M_i = \log(1 + \hat{V}_i) - \log(1 + V_i).
\end{equation}
This relative view aligns with the scouting heuristics. To create actionable targets, we flag a player as \emph{undervalued} if their mispricing exceeds the 85th percentile:
\begin{equation}
u_i = \mathbf{1}(M_i \geq \tau) \quad \text{where} \quad \tau = Q_{0.85}(M).
\end{equation}
This 15\% threshold reflects a realistic \emph{shortlisting budget} in scouting funnels. This relative view aligns with scouting heuristics, where data serve as a high-volume filter to manage limited human observation capacity~\cite{hintz2022moneyball}. Sensitivity analyses at 10\% and 20\% preserved model rankings, suggesting stability.

\subsubsection{Shortlisting Model.}
We trained classifiers (XGBoost, Random Forest, TabNet) to predict undervaluation labels using the full feature set, applying class weighting to address imbalance. Because the label is derived indirectly, we emphasize \emph{ranking quality} via ROC-AUC, treating the classifier as a \emph{shortlister} for human evaluation rather than a hard decision rule. F1-score is reported for completeness. Formally, a classifier estimates:
\begin{equation}
P(U_i = 1 \mid \mathbf{x}_i,\mathbf{z}_i).
\end{equation}

\subsection{Evaluation Protocol}
All experiments used a chronological split: earliest 80\% for training, latest 20\% for testing. PCA and normalization were fitted on training data only. For regression, we compared linear regression, XGBoost, and TabNet. XGBoost hyperparameters: 500 trees, learning rate 0.05, max depth 6, subsampling 0.8. TabNet used early stopping on validation RMSE. Ablation experiments evaluated feature groups: full, no text, and text-only.

\section{Results}
\subsection{Expected Market Value Estimation}
The first stage of our pipeline involves predicting the expected log-market value, against which current value is compared to derive the mispricing label. Table~\ref{tab:regression_comparison} summarizes the performance of the three regression models on the held-out test set: XGBoost, TabNet and Linear Regression.

The gradient-boosted model (XGBoost) provides the strongest fit, explaining 0.935 of the variance in the log-transformed target. This significant performance edge over the linear baseline (R$^2$=0.611) suggests that the relationship between player attributes, news narratives and market value is inherently non-linear and benefits from the ensemble's ability to model complex feature interactions. TabNet, while outperforming the linear model, lags behind XGBoost, suggesting that for this well-structured tabular task with limited sample size, tree-based methods remain more effective than deep learning alternatives.

\begin{figure}[tbh]
\centering
\includegraphics[width=0.70\textwidth]{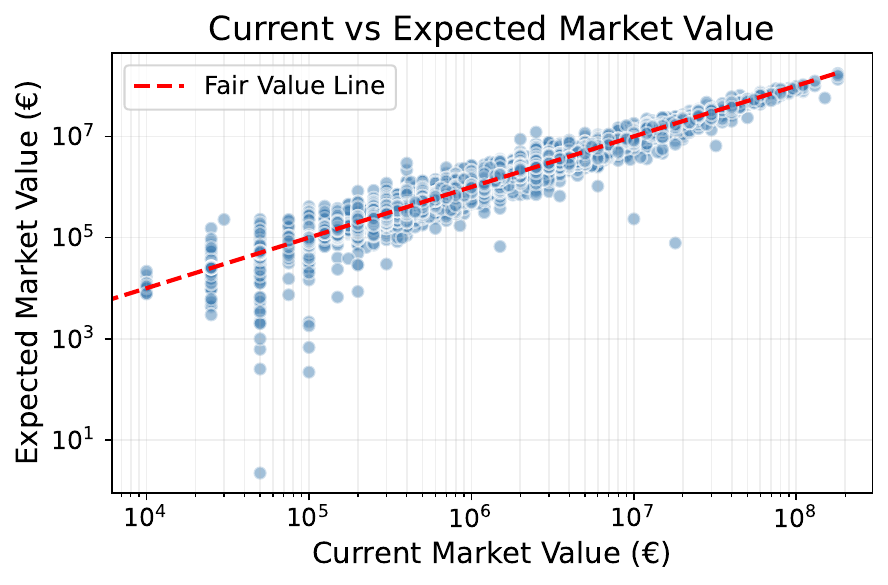}
\caption{Observed versus expected market value (log scale). Players above the diagonal ($y > x$) exhibit positive mispricing and are shortlisted as undervalued. The spread of points illustrates the model's ability to identify valuation discrepancies.}
\label{fig:mispricing}
\end{figure}

Figure~\ref{fig:mispricing} visualizes the relationship between a player's observed market value and the value predicted by the XGBoost model. Points lying on or near the diagonal represent players whose current market value closely aligns with the model's expectation, i.e., they are considered fairly priced. Crucially, the points above the line are those where the predicted value exceeds the observed value, flagging them as potentially \textit{undervalued} and therefore of interest for scouting. Conversely, points below the line indicate potentially \textit{overvalued} players. This visualization serves as an intuitive screening tool for analysts, translating a complex regression output into a clear, actionable shortlist.

\begin{table}[tbh]
\centering
\caption{Regression model performance for expected log-market value. XGBoost substantially outperforms both the neural and linear baselines, capturing the complex and non-linear relationships in the data.}
\label{tab:regression_comparison}
\begin{tabular}{lccc}
\toprule
\textbf{Model} & \textbf{RMSE} $\downarrow$ & \textbf{MAE} $\downarrow$ & \textbf{R$^2$} $\uparrow$ \\
\midrule
XGBoost & \textbf{0.453} & \textbf{0.337} & \textbf{0.935} \\
TabNet     & 0.920 & 0.695 & 0.732 \\
Linear Regression & 1.109 & 0.863 & 0.611 \\
\bottomrule
\end{tabular}
\end{table}

\subsection{Shortlisting Performance and Feature Ablation}
The core contribution of this work is the classification model that shortlists undervalued players. To understand the contribution of different data modalities, we conducted an ablation study, comparing model performance when trained on the full feature set (Full), on market/biographical features only (No Text), and on NLP features only (Text Only). Results are presented in Table~\ref{tab:ablation} with three classification models: XGBoost, TabNet, and Random Forest (RF).

\begin{table}[tbh]
\centering
\caption{Ablation study for the shortlisting task (undervalued vs. non-undervalued). Performance is measured by ROC-AUC, which emphasizes ranking quality. The full model, incorporating both market and text features, consistently achieves the highest discriminative power.}
\label{tab:ablation}
\begin{tabular}{lcccccc}
\toprule
& \multicolumn{2}{c}{XGBoost} & \multicolumn{2}{c}{TabNet} & \multicolumn{2}{c}{Random Forest (RF)} \\
\cmidrule(r){2-3} \cmidrule(lr){4-5} \cmidrule(l){6-7}
Variant & ROC AUC & F1 & ROC AUC & F1 & ROC AUC & F1 \\
\midrule
\rowcolor{blue!5} Full Model & \textbf{0.677} & \textbf{0.263} & 0.614 & 0.206 & 0.471 & 0.137 \\
\rowcolor{white} No Text    & \textbf{0.604} & \textbf{0.250} & 0.560 & 0.247 & 0.553 & 0.208 \\
\rowcolor{blue!5} Text Only  & \textbf{0.514} & \textbf{0.264} & 0.508 & 0.295 & 0.488 & 0.050 \\
\bottomrule
\end{tabular}
\end{table}

Several key observations emerge from this analysis. First, the full model, which combines structured market data with NLP-derived news features, achieves the highest ranking performance (ROC-AUC of 0.677 with XGBoost). This suggests our core hypothesis that media narratives provide a valuable and complementary signal for identifying valuation discrepancies.

Second, removing the textual features (No Text) results in a small but systematic decline in ROC-AUC across all model families. This consistent drop indicates that while market dynamics are the dominant driver of value, the information encoded in news coverage (i.e., sentiment, player form and injury news) offers a genuine and non-redundant improvement to the model's discriminative ability. This aligns with domain knowledge that narratives can contextualize or precede market adjustments.

Third, \textit{Text Only} models perform only marginally above random (ROC-AUC $\approx$ 0.51). This finding is equally important: it demonstrates that while news narratives are a useful supplementary signal, they are insufficient on their own to explain market mispricing. Text provides context to market data, but does not replace it. This result underscores the necessity of a multimodal approach.

Finally, XGBoost consistently outperforms TabNet and Random Forest in the full-model setting. Its higher F1 score also suggests it is better at balancing precision and recall for the minority (\textit{undervalued}) class. This, combined with its regression performance, establishes XGBoost as the more suitable model for this task, offering a strong balance of predictive power, interpretability, and practical deployability.

\begin{figure}[tbh]
\centering
\includegraphics[width=0.60\textwidth]{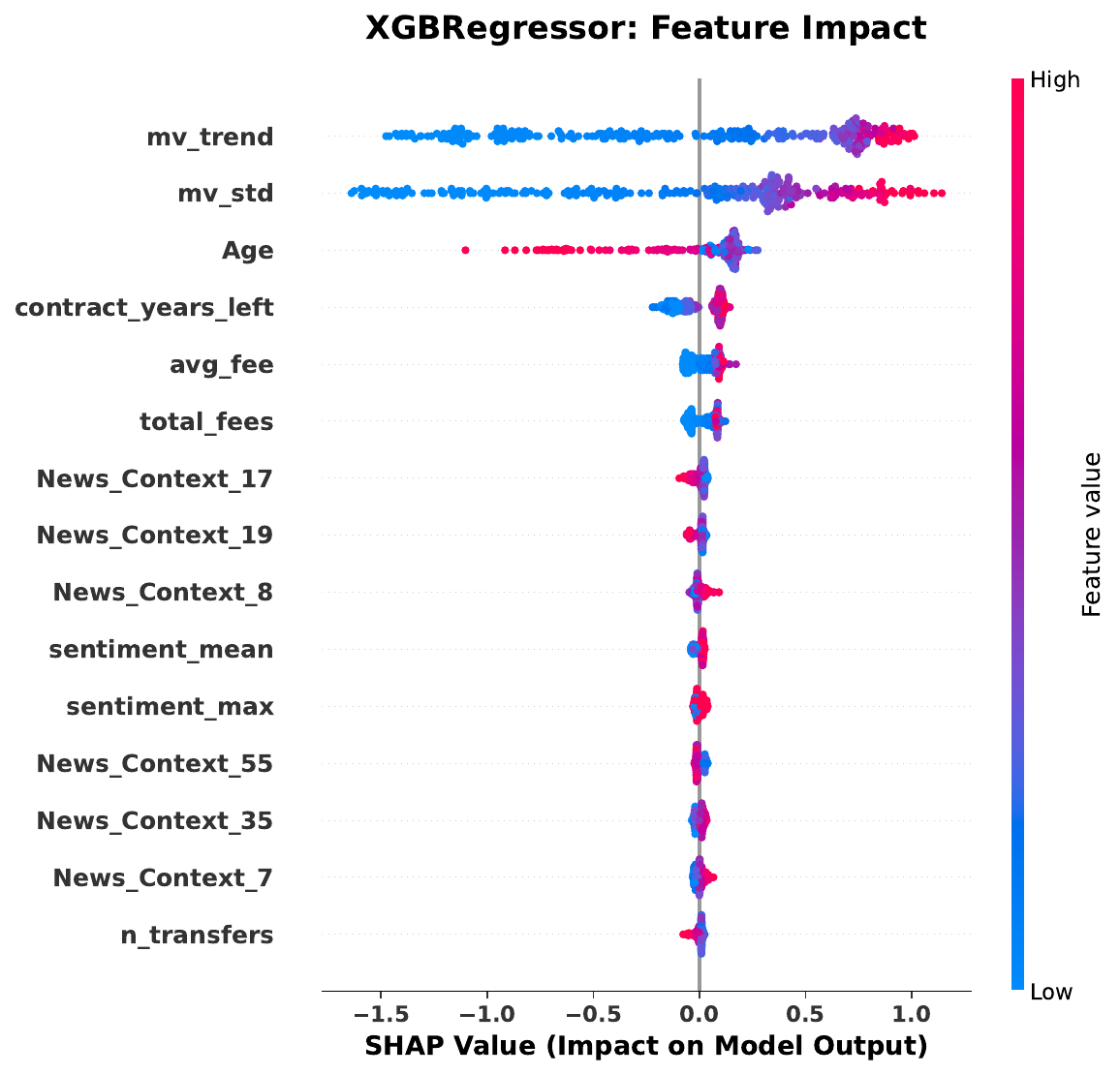}
\caption{SHAP Summary Plot for the XGBoost Regression Model, showing each feature’s contribution to predicted log‑market value. Features are ranked by their global impact, with red indicating high feature values and blue indicating low values.}
\label{fig:Shap}
\end{figure}

\subsection{Model Interpretation}
To move beyond performance metrics and understand the decision-making logic of the regressor, we analyzed the trained XGBoost model using SHAP\footnote{\url{https://shap.readthedocs.io/en/latest/}} (SHapley Additive exPlanations). The SHAP analysis (Figure~\ref{fig:Shap}) indicates that the XGBoost regressor relies primarily on temporal market‑value signals, with \texttt{mv\_trend} and \texttt{mv\_std} exerting the largest influence on predictions. Upward value trajectories strongly increase estimated market value, whereas higher volatility leads to conservative predictions. Age is the next most influential factor, with older ages contributing negatively in line with established valuation dynamics. Contract‑related variables and historical fee indicators show moderate but consistent effects, reflecting their relevance to long‑term value retention. In contrast, the news‑derived contextual features and sentiment measures have only marginal, case‑specific contributions, suggesting that textual information provides supplementary rather than structural predictive value. Overall, the model’s behavior is dominated by market‑based and demographic variables, with contextual signals playing a secondary role.

\section{Deployment Considerations}

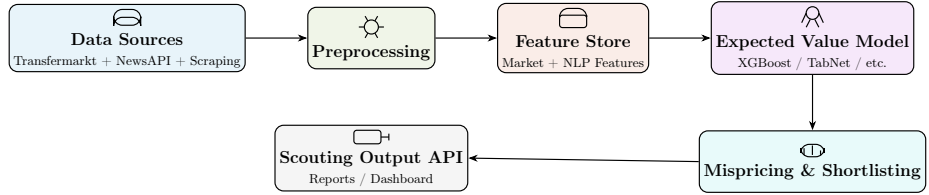
\begin{figure}[t]
\centering
\resizebox{\linewidth}{!}{%
\begin{tikzpicture}[
    node distance=1.5cm,
    box/.style={
        rectangle,
        rounded corners=6pt,
        minimum width=2.5cm,
        minimum height=1.5cm,
        align=center,
        font=\large,
        draw=black,
        thick
    },
    arrow/.style={->, thick, >=Stealth}
]

\definecolor{boxblue}{RGB}{232,244,250}
\definecolor{boxgreen}{RGB}{240,245,232}
\definecolor{boxorange}{RGB}{250,237,232}
\definecolor{boxpurple}{RGB}{246,232,250}
\definecolor{boxcyan}{RGB}{232,250,250}
\definecolor{boxgrey}{RGB}{245,245,245}

\newcommand{\iconData}{%
  \begin{tikzpicture}[scale=0.55]
    \draw[thick] (0,0) -- (1.2,0) -- (1.2,0.35) -- (0,0.35) -- cycle;
    \draw[thick] (0,0.35) -- (0,0.7) -- (1.2,0.7) -- (1.2,0.35);
  \end{tikzpicture}%
}
\newcommand{\iconGear}{%
  \begin{tikzpicture}[scale=0.45]
    \draw[thick] (0,0) circle (0.38);
    \foreach \x in {0,60,...,300} \draw[thick] (\x:0.38) -- (\x:0.68);
  \end{tikzpicture}%
}
\newcommand{\iconFolder}{%
  \begin{tikzpicture}[scale=0.55]
    \draw[thick, rounded corners=2pt] (0,0) rectangle (1.2,0.55);
    \draw[thick] (0,0.55) -- (0,0.85) -- (1.2,0.85) -- (1.2,0.55);
  \end{tikzpicture}%
}
\newcommand{\iconModel}{%
  \begin{tikzpicture}[scale=0.55]
    \draw[thick] (0,0.28) circle (0.28);
    \draw[thick] (-0.28,0.28) -- (-0.48,-0.15);
    \draw[thick] (0.28,0.28) -- (0.48,-0.15);
    \draw[thick] (-0.1,0) -- (-0.28,-0.35);
    \draw[thick] (0.1,0) -- (0.28,-0.35);
  \end{tikzpicture}%
}
\newcommand{\iconBars}{%
  \begin{tikzpicture}[scale=0.55]
    \draw[thick] (0,0) rectangle (1.05,0.55);
    \draw[thick] (0.15,0.08) -- (0.15,0.38);
    \draw[thick] (0.52,0.08) -- (0.52,0.50);
    \draw[thick] (0.88,0.08) -- (0.88,0.28);
  \end{tikzpicture}%
}
\newcommand{\iconAPI}{%
  \begin{tikzpicture}[scale=0.55]
    \draw[thick, rounded corners=2pt] (0,0) rectangle (1.2,0.6);
    \draw[thick] (1.2,0.3) -- (1.55,0.3);
    \draw[thick] (1.55,0.15) -- (1.55,0.45);
  \end{tikzpicture}%
}

\node[box, fill=boxblue] (data) {\iconData\\[-1pt]\textbf{Data Sources}\\ \small Transfermarkt + NewsAPI + Scraping};
\node[box, fill=boxgreen, right=of data] (pre) {\iconGear\\[-1pt]\textbf{Preprocessing}};
\node[box, fill=boxorange, right=of pre] (store) {\iconFolder\\[-1pt]\textbf{Feature Store}\\ \small Market + NLP Features};
\node[box, fill=boxpurple, right=of store] (value) {\iconModel\\[-1pt]\textbf{Expected Value Model}\\ \small XGBoost / TabNet / etc.};

\node[box, fill=boxgrey, below=1.35cm of pre, xshift=0.0cm] (api) {\iconAPI\\[-1pt]\textbf{Scouting Output API}\\ \small Reports / Dashboard};
\node[box, fill=boxcyan, below=1.35cm of value, xshift=0.0cm] (mis) {\iconBars\\[-1pt]\textbf{Mispricing \& Shortlisting}};

\draw[arrow] (data) -- (pre);
\draw[arrow] (pre) -- (store);
\draw[arrow] (store) -- (value);
\draw[arrow] (value.south) -- (mis.north);
\draw[arrow] (mis.west) -- (api.east); 

\end{tikzpicture}%
}
\caption{Prototype deployment architecture for the multimodal undervaluation shortlisting system.}
\label{fig:deployment-pipeline}
\end{figure}

To illustrate how the proposed framework can be incorporated into real scouting workflows, Figure~\ref{fig:deployment-pipeline} presents a prototype deployment architecture assembled around the multimodal valuation pipeline. The system ingests weekly Transfermarkt market-value updates alongside continuously collected football news articles retrieved via NewsAPI and compliant web scraping. These inputs are processed through a unified preprocessing component that performs temporal alignment, entity normalisation, market-dynamics computation, and text cleaning. The resulting structured and textual features are stored in a lightweight feature repository.

At each update cycle, the expected-value model is executed to produce revised predictions of players’ market valuations. Mispricing scores are then computed by comparing the predicted and observed values, after which the shortlisting module ranks players based on the mispricing threshold. This produces an updated candidate list that can be used by downstream analytics tools or scouting staff.

To integrate with existing decision-support systems, the output is served via an internal API that exposes shortlists, player-level explanations, and mispricing trajectories. The entire workflow is computationally efficient and can be scheduled on a single GPU-enabled workstation or mid-tier server, enabling weekly or daily refresh intervals. The deployment pipeline demonstrates that the proposed methodology is not only analytically sound but also technically feasible for near-real-time operational use in professional scouting environments.

\section{Discussion}
The experimental results demonstrate that combining structured market information with news-derived textual representations provides a practical and interpretable approach for identifying players whose market valuations deviate from model-based expectations. Market dynamics remain the primary drivers of valuation, confirming economic intuition that long-term trends, volatility, and career stability carry the strongest explanatory power. News-derived features, although not dominant in isolation, consistently provide complementary value by capturing evolving narratives related to form, injury updates, tactical adjustments, and transfer speculation. These factors are often qualitatively expressed in news coverage before they manifest in observable market movements.

Importantly, the mispricing-based formulation aligns well with real scouting workflows, where analysts rarely rely on a single metric but instead prioritize shortlisting candidates for deeper review. Treating undervaluation prediction as a ranking task rather than a strict binary classification mirrors this practice and enables stable identification of high-potential targets under realistic class imbalance. The multimodal framework also offers interpretability benefits: sentiment volatility and semantic embedding components suggest when narrative uncertainty may coincide with market underreaction, potentially revealing overlooked opportunities. These findings collectively indicate that textual information does not replace structured market data, but serves as a robust supplementary signal that improves shortlisting reliability in complex, real-time decision-making environments.

\section{Limitations and Future Work}
Despite its practical value, the framework has several limitations. First, Transfermarkt valuations (though widely used) are proxies that may contain platform-specific biases, crowd effects, or noise unrelated to true economic value. Second, the news corpus predominantly reflects English-language coverage, concentrating on leagues with higher media visibility and potentially overlooking narratives surrounding lower-profile markets. Third, while chronological splitting prevents information leakage, the framework remains observational and does not establish causality; textual signals may reflect market dynamics rather than precede them.

Future work will address these challenges by modeling lag structures to quantify whether news sentiment and narratives lead or follow changes in market valuation, extending experiments across leagues with diverse media ecosystems, and incorporating multilingual or region-specific news sources. Causal inference methods may help assess whether particular narrative themes systematically precede mispricing corrections. In addition, more sophisticated aggregation mechanisms (e.g., temporal decay, source weighting, attention-based document pooling) could enhance the extraction of semantically influential articles. Finally, deployment-oriented evaluation involving Precision@$k$, calibration analysis, and user studies with practitioners would deepen understanding of how the system performs in operational scouting environments.

\section{Conclusion}
This work introduced a multimodal, leakage-aware framework for detecting undervalued football players through an objective mispricing signal that compares model-estimated and market-observed valuations. By integrating structured market dynamics with sentiment and semantic features extracted from football news, the framework provides a practical shortlisting tool that aligns with real-world scouting processes. Market-driven features dominate predictive performance, while news-derived features consistently add complementary contextual information that enhances robustness and interpretability.

The approach is transparent, readily extensible, and operationally feasible for real-time or periodic valuation updates. It offers analysts a reproducible method to surface potentially overlooked players and highlights the value of combining quantitative signals with narrative-driven textual context. As football markets continue to evolve and media ecosystems diversify, such multimodal, data-driven strategies represent a promising direction for next-generation scouting analytics.

\section{Reproducibility and Ethics}
All experiments were conducted using a strict chronological evaluation protocol, ensuring that no future information was available to training procedures. Preprocessing operations, including PCA dimensionality reduction and text normalization, were fitted exclusively on the training split and applied unchanged to validation and test data. Random seeds were fixed for all model training and feature transformations to ensure deterministic behavior. Experiments were executed on a single CUDA-enabled GPU and commodity CPU hardware; detailed specifications and configuration files will be provided in the accompanying code repository, subject to licensing constraints on underlying data sources.

Structured player data were collected from Transfermarkt, while the news corpus was constructed from publicly accessible sources via NewsAPI and ethically compliant web scraping. Only publicly published information was processed, and no sensitive or private player data were included. Redistribution of raw news or structured datasets is restricted by source licensing; however, derived features and full preprocessing code will be released upon acceptance. The system is intended strictly as a decision-support tool. Analysts and clubs should avoid overreliance on automated outputs without expert oversight, particularly when evaluating players in contexts where media coverage may be uneven or biased.

\bibliographystyle{splncs04}
\bibliography{mybibliography}

@article{franck2012talent,
  title={Talent and/or popularity: what does it take to be a superstar?},
  author={Franck, Egon and N{\"u}esch, Stephan},
  journal={Economic Inquiry},
  volume={50},
  number={1},
  pages={202--216},
  year={2012},
  publisher={Wiley Online Library}
}

@article{herm2014crowd,
  title={When the crowd evaluates soccer players’ market values: Accuracy and evaluation attributes of an online community},
  author={Herm, Steffen and Callsen-Bracker, Hans-Markus and Kreis, Henning},
  journal={Sport Management Review},
  volume={17},
  number={4},
  pages={484--492},
  year={2014},
  publisher={Elsevier}
}

@article{franceschi2024determinants,
  title={Determinants of football players’ valuation: A systematic review},
  author={Franceschi, Maxence and Brocard, Jean-Fran{\c{c}}ois and Follert, Florian and Gouguet, Jean-Jacques},
  journal={Journal of Economic Surveys},
  volume={38},
  number={3},
  pages={577--600},
  year={2024},
  publisher={Wiley Online Library}
}

@article{van2011valuation,
  title={The valuation of human capital in the football player transfer market},
  author={Van den Berg, Erik},
  journal={Rotterdam: ErasmusUniversity},
  year={2011}
}

@article{tetlock2007giving,
  title={Giving content to investor sentiment: The role of media in the stock market},
  author={Tetlock, Paul C},
  journal={The Journal of finance},
  volume={62},
  number={3},
  pages={1139--1168},
  year={2007},
  publisher={Wiley Online Library}


}

@inproceedings{decroos2019actions,
  title={Actions speak louder than goals: Valuing player actions in soccer},
  author={Decroos, Tom and Bransen, Lotte and Van Haaren, Jan and Davis, Jesse},
  booktitle={Proceedings of the 25th ACM SIGKDD international conference on knowledge discovery \& data mining},
  pages={1851--1861},
  year={2019}
}

@article{bollen2011twitter,
  title={Twitter mood predicts the stock market},
  author={Bollen, Johan and Mao, Huina and Zeng, Xiaojun},
  journal={Journal of computational science},
  volume={2},
  number={1},
  pages={1--8},
  year={2011},
  publisher={Elsevier}
}

@article{megia2025football,
  title={Football Fan Satisfaction Based on Ticket Price. A Dynamic Hedonic Pricing Model Approach},
  author={Meg{\'\i}a-Cayuela, Daniel},
  journal={Social Science Quarterly},
  volume={106},
  number={5},
  pages={e70071},
  year={2025},
  publisher={Wiley Online Library}
}

@article{behravan2021novel,
  title={A novel machine learning method for estimating football players’ value in the transfer market},
  author={Behravan, Iman and Razavi, Seyed Mohammad},
  journal={Soft Computing},
  volume={25},
  number={3},
  pages={2499--2511},
  year={2021},
  publisher={Springer}
}

@article{schumaker2016predicting,
  title={Predicting wins and spread in the Premier League using a sentiment analysis of twitter},
  author={Schumaker, Robert P and Jarmoszko, A Tomasz and Labedz Jr, Chester S},
  journal={Decision Support Systems},
  volume={88},
  pages={76--84},
  year={2016},
  publisher={Elsevier}



}

@article{h2021quantifying,
  title={Quantifying fan engagement in sports using text analytics},
  author={H. Zadeh, Amir},
  journal={Journal of Data, Information and Management},
  volume={3},
  number={3},
  pages={197--208},
  year={2021},
  publisher={Springer}
}

@article{jacobsen2023managing,
  title={Managing institutional complexity in a football organization},
  author={Jacobsen, {\AA}se},
  journal={Managing Sport and Leisure},
  pages={1--20},
  year={2023},
  publisher={Taylor \& Francis}
}

@article{follert2024decision,
  title={A decision model to value football player investments under uncertainty},
  author={Follert, Florian and Glei{\ss}ner, Werner},
  journal={Management Decision},
  volume={62},
  number={13},
  pages={178--200},
  year={2024},
  publisher={Emerald Publishing Limited}
}

@article{waskita2025role,
  title={The role of machine learning in modern football analytics: A systematic review of approaches and their implications},
  author={Waskita, Ghozi Indra and Kurniawan, Haris and Yudhistira, Dewangga and Mohamad, Nur Ikhwan Bin and Anam, M Khairul},
  journal={Journal of Sport and Exercise Science},
  volume={8},
  number={2},
  pages={178--186},
  year={2025}
}

@article{mammadov2024optimizing,
  title={OPT{\.I}M{\.I}Z{\.I}NG QUERY UNDERSTAND{\.I}NG: A MATHEMAT{\.I}CAL APPROACH TO NLP IN SEARCH ENG{\.I}NES},
  author={Mammadov, E},
  journal={Norwegian Journal of development of the International Sci ence No},
  volume={129},
  pages={91},
  year={2024}
}

@article{davidovic2025news,
  title={News sentiment and stock market dynamics: A machine learning investigation},
  author={Davidovic, Milivoje and McCleary, Jacqueline},
  journal={Journal of Risk and Financial Management},
  volume={18},
  number={8},
  pages={412},
  year={2025},
  publisher={MDPI}
}

@inproceedings{xia2023mole,
  title={Mole-bert: Rethinking pre-training graph neural networks for molecules},
  author={Xia, Jun and Zhao, Chengshuai and Hu, Bozhen and Gao, Zhangyang and Tan, Cheng and Liu, Yue and Li, Siyuan and Li, Stan Z},
  booktitle={The Eleventh International Conference on Learning Representations},
  year={2023}
}

@article{reimers2019sentence,
  title={Sentence-bert: Sentence embeddings using siamese bert-networks},
  author={Reimers, Nils and Gurevych, Iryna},
  journal={arXiv preprint arXiv:1908.10084},
  year={2019}
}

@article{vaswani2017attention,
  title={Attention is all you need},
  author={Vaswani, Ashish and Shazeer, Noam and Parmar, Niki and Uszkoreit, Jakob and Jones, Llion and Gomez, Aidan N and Kaiser, {\L}ukasz and Polosukhin, Illia},
  journal={Advances in neural information processing systems},
  volume={30},
  year={2017}
}

@inproceedings{devlin2019bert,
  title={Bert: Pre-training of deep bidirectional transformers for language understanding},
  author={Devlin, Jacob and Chang, Ming-Wei and Lee, Kenton and Toutanova, Kristina},
  booktitle={Proceedings of the 2019 conference of the North American chapter of the association for computational linguistics: human language technologies, volume 1 (long and short papers)},
  pages={4171--4186},
  year={2019}
}

@article{sanh2019distilbert,
  title={DistilBERT, a distilled version of BERT: smaller, faster, cheaper and lighter},
  author={Sanh, Victor and Debut, Lysandre and Chaumond, Julien and Wolf, Thomas},
  journal={arXiv preprint arXiv:1910.01108},
  year={2019}
}

@manual{spacy,
  title  = {{spaCy: Industrial-strength Natural Language Processing in Python}},
  author = {Honnibal, Matthew and Montani, Ines},
  year   = {2020},
  url    = {https://spacy.io/}
}

@article{mccarthy2007vocd,
  title={vocd: A theoretical and empirical evaluation},
  author={McCarthy, Philip M and Jarvis, Scott},
  journal={Language Testing},
  volume={24},
  number={4},
  pages={459--488},
  year={2007},
  publisher={Sage Publications Sage UK: London, England}
}

@inproceedings{hintz2022moneyball,
  title={Moneyball: The computational turn in professional sports management},
  author={Hintz, Eric S},
  booktitle={Papers of the Business History Conference},
  year={2022}
}

\end{document}